\documentclass[a4paper,11pt]{article}%
\usepackage{listings}
\usepackage{placeins}
\usepackage{multirow}
\usepackage{array}
\usepackage{amssymb}
\usepackage{graphicx}
\usepackage{rotating}
\usepackage{amsfonts}
\usepackage{mathrsfs}
\usepackage{amsmath}
\usepackage{babel}
\usepackage{lastpage}
\usepackage[utf8]{inputenc}
\usepackage[legalpaper,bookmarks=true,colorlinks=true,linkcolor=blue,citecolor=blue]{hyperref}
\usepackage{graphicx}%
\usepackage{fancyhdr,enumerate}
\usepackage{color}
\usepackage[mathlines]{lineno}
\usepackage{lscape}
\usepackage{epsfig}
\usepackage[sort&compress]{natbib}
\usepackage{epstopdf}% To incorporate .eps illustrations using PDFLaTeX, etc.
\usepackage{subfigure}
\usepackage{float,adjustbox}
\usepackage{bm}
\usepackage{booktabs}
\usepackage{geometry}
\usepackage{geometry}
\begin{document}
\title{A Comparative Analysis of Traditional and Deep Learning Time Series Architectures for Influenza \textbf{A} Infectious Disease Forecasting}

\author{Edmund F. Agyemang$^{\textbf{1,2*}}$, Hansapani Rodrigo$^{\textbf{1*}}$ \& Vincent Agbenyeavu$^{\textbf{1}}$\\ $^{\textbf{1}}$	School of Mathematical and Statistical Sciences,\\ College of Sciences, University of Texas Rio Grande Valley, USA.\\
 $^{\textbf{2}}$Department of Biostatistics and Data Science, Celia Scott Weatherhead School of \\ Public Health and Tropical Medicine at Tulane University, New Orleans, Louisiana, USA\\	\vspace{0.1in}  $^{\textbf{*}}$Author to whom any correspondence should be addressed\\ \vspace{0.1in} 
 Email: edmundfosu6@gmail.com, \\ 
 \noindent Keywords: Influenza \textbf{A}, ARIMA, ETS, GRU, LSTM, Transformer}

\date{}
\maketitle
\begin{abstract}% Abstract in English.\medskip
	\noindent	
%done
Influenza \textbf{A} is responsible for 290,000 to 650,000 respiratory deaths a year, though this estimate is an improvement from years past due to improvements in sanitation, healthcare practices, and vaccination programs. In this study, we perform a comparative analysis of traditional and deep learning models to predict Influenza \textbf{A} outbreaks. Using historical data from January 2009 to December 2023, we compared the performance of traditional ARIMA and Exponential Smoothing (ETS) models with six distinct deep learning architectures: Simple RNN, LSTM, GRU, BiLSTM, BiGRU, and Transformer. The results reveal a clear superiority of all the deep learning models, especially the state-of-the-art Transformer with respective average testing MSE and MAE of $0.0433 \pm 0.0020$ and $0.1126 \pm 0.0016$ for capturing the temporal complexities associated with Influenza \textbf{A} data, outperforming well known traditional baseline ARIMA and ETS models. These findings of this study provide evidence that state-of-the-art deep learning architectures can enhance predictive modeling for infectious diseases and indicate a more general trend toward using deep learning methods to enhance public health forecasting and intervention planning strategies.  Future work should focus on how these models can be incorporated into real-time forecasting and preparedness systems at an epidemic level, and integrated into existing surveillance systems.
\\
\end{abstract}

\section{Introduction}
%done
Influenza and COVID-19 regarded as some of the deadliest respiratory pandemics seen in history, have devastated populations across the globe \citep{choreno2021clinical, nyarko2023covid}. The pathogens responsible for these diseases have changed and mutated over time and continue to do so, with serious consequences for how well we can predict, diagnose, and understand them. Worst-case scenarios are sobering: What if the pathogens causing Influenza \textbf{A} mutated significantly, causing even deadlier and more contagious forms of the diseases?. This makes it a worrying concern for public health authorities \citep{kim2018influenza}. Likewise, both pathogens have demonstrated potential for global spread via an interconnected web of modern transportation systems. This global spread is made worse by the times when infected people display no symptoms and so cannot be known to be sick. These are the ``asymptomatic transmission phases" that must be figured into predictive models to understand the full picture of an Influenza A infectious disease's behavior. These challenges highlight the need for strong surveillance systems, extensive research, and unprecedented global collaboration among virologists, epidemiologists, and vaccine developers. These diseases have not only health-related effects but also substantial effects on economies, societies, and daily lives worldwide \citep{bloom2022modern}. In the United States, the  Centers for Disease Control and Prevention (CDC), an Influenza like illness (ILI) surveillance network, publishes weekly (CDC weekly ILI reports) based on its surveillance networks. Although vital in tracking disease burden, typically, there is at least a 1-week delay of ILI reports due to data collation and aggregation. An unusual case: CDC did not report data for 2020-2021 Influenza cases. Given this delay, forecasting Influenza \textbf{A} activity is essential for real time disease surveillance and for public health agencies to direct resources to prepare and plan for potential pandemics. It is in this regard that this study comes in handy to fill the gap by comparing the forecasting power of competing traditional time series models [Auto Regressive Integrated Moving Averages (ARIMA) and Holt-Winters Exponential Smoothing (ETS)], and state-of-the-art deep learning time series models [Simple Recurrent Neural Network (RNN), Long Short Term Memory (LSTM), Gated Recurrent Unit (GRU), BiLSTM, BiGRU, and Transformer] in analyzing Influenza \textbf{A} dynamics and to accurately forecast ahead of time.
\\

\noindent
%done
Seasonal Influenza causes around 400,000 deaths worldwide annually as a result of respiratory diseases. During major flu pandemics caused by significant mutations in Influenza strains, this number can increase drastically \citep{paget2019global}. Despite this grim statistic, Influenza mortality has dramatically reduced over the years, thanks to better sanitation, healthcare infrastructure, and cost-effective vaccination. At the same time, despite these improvements, Influenza remains a major global health threat, worsened by factors such as an aging global population, lack of access to healthcare systems, and poor social conditions in many parts of the world.  The death rate associated with Influenza \textbf{A} has fallen significantly over the years, a trend evident from analyzing historical U.S.A data to understand “cohort effects." This shows that recent generations, when controlling for age, are less likely to die of Influenza \textbf{A} than earlier generations \citep{hansen2022mortality}. There has been a gradual decline in flu-related deaths since the start of the 20th century. This may be due to various reasons, notably the introduction of widespread sanitation programs in American cities in the early 20th century \citep{cutler2005role}. And in the 20th century, advances in neonatal care and increased childhood immunization reduced comorbidities that heighten Influenza \textbf{A}'s mortality risks. Another important reason has been the development and greater use of vaccines against the flu virus. Over the years, the vaccination levels have gone up, especially among older population.
 \\

\noindent
%done
Numerous statistical and mathematical approaches have been employed in the analysis of Influenza \textbf{A} cases. Among them, the SIR (Susceptible, Infected, Recovered) and SEIR (Susceptible, Exposed, Infected, Recovered) models are well known and are discussed in several contexts, including the context of Influenza \textbf{A} epidemics, famously discussed in \cite{coburn2009modeling}. The study also emphasizes the impact of various public health measures implemented during Influenza \textbf{A} epidemics. These measures include vaccination, timely organization of quarantine centers, and social distancing.  The study also assesses the feasibility of these and other measures in the same breath and under the same context required to prevent Influenza \textbf{A} epidemic. \cite{beauchemin2011review},  however, focus on the within-host dynamics of Influenza \textbf{A} adopting basic mathematical frameworks, without immune response, and their applications. These models, which consist mainly of ordinary differential equations (ODEs), monitor the dynamics between susceptible cells, infected cells, and free virions. Even though the models are simple, they do provide insight into infection kinetics, and they expose the limitations of assuming exponentially distributed latent and infectious periods. Interestingly, the study also discusses higher order models  (models that are better at approximately simulating in vivo dynamics) that are more in line with the infection process once one takes the immune response into consideration.
  \\

\noindent
%done
According to \cite{boianelli2015modeling}, due to seasonal epidemics and pandemics, Influenza \textbf{A} virus (IAV) infections pose a global threat. The study highlights the difficulty of understanding the interaction between the virus and the host immune response. The study also discusses the use of mathematical modeling to explain and quantify IAV dynamics and provides insights into the control of the adaptive immune response and the factors that support the development of secondary bacterial co-infection. Transmission of Influenza A occurs when infectious and susceptible individuals come into contact, according to modeling studies. However, not all contacts result in infection, and the ones that do are determined by two main factors: how infectious the infected person is and how susceptible the contact is \citep{chen2025global}. Therefore, one must first know the background of Influenza \textbf{A} and its possibility for producing worldwide pandemics if one is to create suitable public health responses and get ready for next outbreaks. This capacity of the virus demands careful study on its genetic variants, dynamics of transmission, and treatment strategies \citep{lin2008rapid}.\\

\noindent
The objective of this study is to conduct a comprehensive analysis of traditional statistical models (ARIMA and ETS) and state-of-the-art deep learning models (RNN, LSTM, GRU, BiLSTM, BiGRU, and Transformer) to assess their applicability for reliable and accurate Influenza \textbf{A} dynamics forecasting. This comparative analysis aims to support the creation of increasingly dependable and precise models that forecast the likelihood of an influenza \textbf{A} outbreak occurring at some future date. The hope is that such models will be useful in the development of public health measures that serve to reduce the impact of Influenza \textbf{A} outbreaks when they do occur.  The remainder of the paper is organized as follows:  Section \ref{Sec2} discusses the data and methods used for the study. Section \ref{Sec3} provides the mathematical framework of the study. Section \ref{Sec4} discusses the results and findings of the study whilst section \ref{Sec5} concludes the study and provide recommendations for further work.

 \section{Data and Methods \label{Sec2}}
 Secondary data on Influenza \textbf{A} cases which consists of variants such as  H1N12009 and H7N9 from January 2009 to December 2023 were retrieved from Our World in Data and can be accessed at \url{https://ourworldindata.org/influenza}. The data was analyzed using traditional  ARIMA, ETS time series models, three variants of uni-directional recurrent neural networks (RNN) (RNN, LSTM and GRUs), two variants of bi-directional RNN (BiLSTM and BiGRU) and Transformer. The recorded number of monthly Influenza \textbf{A} cases include all types of surveillance of Influenza \textbf{A} cases including A(H1N12009), A(H3N2) and A(H7N9) Infectious Diseases. Monthly Influenza \textbf{A} cases from January 2009 to December 2022 (168 months) were used in training the traditional and deep learning models models, while the monthly Influenza \textbf{A} cases from January 2023 to December 2023 (12 months) were used in testing the accuracy of the eight models under consideration. We tune the hyperparameters of each deep learning model based on their keras packages. The full specification of the deep learning models after hyperparameter tuning is presented in Table \ref{Tab3}. The study's analysis was conducted using the R version 4.5.0 and Python programming languages.

 \section{Methodology \label{Sec3}}
In this section, the methods adopted for the study are discussed.

 \subsection{Simple Recurrent Neural Networks (Simple RNN)}
 \noindent
 RNNs are modifications of feed-forward neural networks with recurrent connections. In a typical neural network, the neuron output at time t is given in (\ref{x}) by:
\begin{equation}
h_t = g(W_iX_t + b)
\label{x}
\end{equation}
In an RNN, the output of the neuron at time $t-1$ is fed back
into the neuron. Central to the simple RNN is its internal memory state, represented as $h_t$.  This state is updated at each time step in accordance with (\ref{Q}):

\begin{equation}
h_t = g(W_iX_t + U_ih_{t-1} + b)
\label{Q}
\end{equation}
\noindent
where $g()$ denotes an activation function, which is selected based on the specific requirements of the task, such as a sigmoid or tanh function. The activation function introduces non-linearity into the model, enabling it to capture more complex patterns in the data. The term $W$ refers to the weight matrix associated with the input vector $X$, and $U$ is the weight matrix for the previous hidden state $h_{t-1}$. The bias term is represented as $b$, playing a crucial role in adjusting the output of the neuron. 

\subsection{Long Short Term Memory (LSTM) Model}
\begin{figure}[!hbt]
	\centering
	\includegraphics[width=0.75\linewidth]{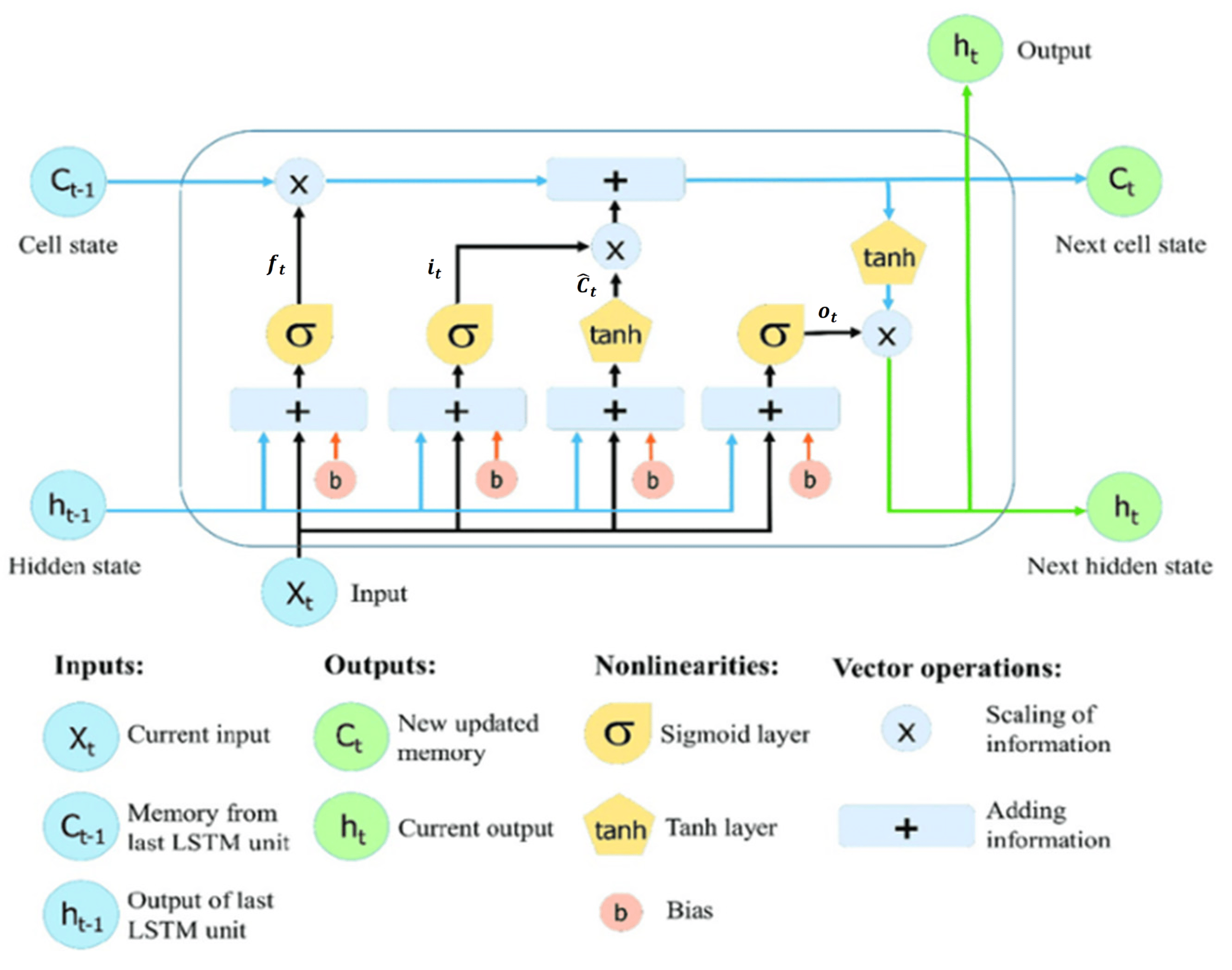}
	\caption{\textbf{Working Architecture of LSTM model adopted from \cite{pierre2023peak}}}
	\label{LSTM}
\end{figure}

\noindent Simple RNNs are known to face challenges when it comes to capturing dependencies that span over long sequences. This is where LSTM networks, an advanced variant of RNNs, come into play, offering a robust solution to the notorious vanishing / gradient issue that plagues simple RNNs. LSTMs with its structure displayed in Figure \ref{LSTM} enhances the RNN architecture with a sophisticated mechanism that significantly extends its memory capacity \cite{hua2019deep}. The gates within an LSTM cell work in concert to maintain the delicate balance
between retaining valuable information and discarding the redundant.  

\noindent
In LSTM, the computation of the output value $h_t$ involves several sequential steps, as defined by the following equations (\ref{Ft}) - (\ref{htt}):
\begin{equation}
	f_t=\sigma\left(W_fX_t+V_fh_{t-1}+b_f\right)
	\label{Ft}
\end{equation}

\begin{equation}
	i_t=\sigma\left(W_iX_t+V_ih_{t-1}+b_i\right)
	\label{it}
\end{equation}

\begin{equation}
	\widetilde{C}_t=\tanh\left(W_cX_t+V_ch_{t-1}+b_c\right)
	\label{ct}
\end{equation}

\begin{equation}
	C_t=f_t \odot C_{t-1}+i_t \odot \widetilde{C}_t
	\label{Ct}
\end{equation}

\begin{equation}
	o_t=\sigma\left(W_oX_t+V_oh_{t-1}+b_o\right)
	\label{ot}
\end{equation}

\begin{equation}
	h_t=o_t \odot \tanh\left(C_t\right)
	\label{htt}
\end{equation}

\noindent
where $W_i$, \(W_f\), $W_o$ and $W_c$ are the weight matrices of the input, forget, output, and memory gates associated with the current input, respectively, whereas $V_i$, $V_f$,$V_o$ and $V_c$ are weight matrices that map the output of the previous hidden state to the corresponding gates of input, forget, output, and memory; $b_i$, $b_f$, $b_o$ and $b_c$ represent the biases of the input, forget, output, and memory gates, respectively. $C_t$ and $h_t$ are the new updated memory and current output, respectively. $\sigma$ and $\tanh$ represent the sigmoid and hyperbolic tangent activation functions, respectively. The operator $\circledcirc$ denotes the Hadamard product (element-wise product).

\subsection{Gated Recurrent Unit (GRU)}
\begin{figure}[!hbt]
	\centering
	\includegraphics[width=0.65\linewidth]{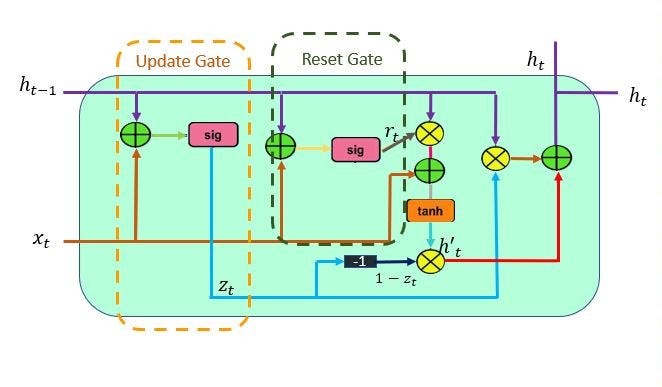}
	\caption{\textbf{Working Architecture of GRU model}}
	\label{GRU}
\end{figure}

\noindent
GRU with its structure displayed in Figure \ref{GRU}, proposed by \cite{cho2014properties}, is a variant of LSTM network. The key modification in GRUs is the amalgamation of the input and forget gates into a unified update gate. This consolidation results in a reduction of the model's parameters compared to the LSTM, thereby facilitating more efficient training.
Likewise, in GRUs, the computation of the output value $h_t$ involves several sequential steps, as defined by the following equations (\ref{rt}) - (\ref{ht}):

\begin{itemize}
		\item Reset Gate:
	\begin{equation}
		r_t = \sigma(W_rX_t + U_rh_{t-1} + b_r)
		\label{rt}
	\end{equation}
	
	\item Update Gate:
	\begin{equation}
		z_t = \sigma(W_zX_t + U_zh_{t-1} + b_z)
		\label{zt}
	\end{equation}

	\item Candidate Hidden State:
	\begin{equation}
		\widetilde{h}_t = \tanh(W_hX_t + U_h(r_t \odot h_{t-1})+b_h)
		\label{hatht}
	\end{equation}
	
	\item Final Output:
	\begin{equation}
		h_t =  z_t \odot h_{t-1} + (1-z_t) \odot \widetilde{h}_t
		\label{ht}
	\end{equation}
\end{itemize}

\noindent
The reset gate $r_t$ determines how much of the previous hidden state (i.e., past information) should be forgotten when calculating the new candidate hidden state, while the update gate $z_t$ is responsible for deciding how much information from the previous hidden state should be retained for the next time step and how much new information from the candidate hidden state should be integrated. This approach differs from the LSTM's forget gate, which offers much better control over memory retention.  In GRUs, the decision to retain or forget past memory is more absolute. Studies have shown that GRUs can match the performance of LSTMs in certain applications, offering similar results with lower computational costs \citep{cho2014properties}. 

\subsection{BiLSTM \& BiGRU}
Let \( \overrightarrow{h_t}=\text{LSTMforward} (x_t,\overrightarrow{h_t}_{-1}) \) and \( \overleftarrow{h_t} =\text{LSTMbackward} (x_t,\overleftarrow{h_t}_{+1}) \). Mathematically, the combined output at a given time step \( t \) in a Bi-LSTM, denoted as \(h_t^{bl}\), is expressed in (\ref{bl}) as:

\begin{equation}
	h_t^{bl} = [\overrightarrow{h_t}, \overleftarrow{h_t}]
	\label{bl}
\end{equation}

\noindent where \( \overrightarrow{h_t}\) and \( \overleftarrow{h_t} \) are the hidden states of the forward and backward LSTMs, respectively at time step $t$, and the square brackets signify concatenation.\\

\noindent
Similarly, Let \( \overrightarrow{h_t}=\text{GRUforward} (x_t,\overrightarrow{h_t}_{-1}) \) and \( \overleftarrow{h_t} =\text{GRUbackward} (x_t,\overleftarrow{h_t}_{+1}) \) be the hidden states of the forward and backward GRU at time step $t$, respectively. The output of the BiGRU at time step $t$ is given in (\ref{tt}) by:

\begin{equation}
	h_t^{bg} = [\overrightarrow{h_t}, \overleftarrow{h_t}]
	\label{tt}
\end{equation}  

\noindent where $[\overrightarrow{h_t}, \overleftarrow{h_t}]$ denotes the concatenation of the forward and backwards hidden states of the GRU network.\\

\noindent
For the mathematical framework of the Transformer model, the reader is directed to \cite{vaswani2017attention}.

\subsection{Exponential Smoothing via Holt-Winters Seasonal Method (ETS)}
Holt-Winters method incorporates seasonality along with level and trend. It can be either additive or multiplicative. For the additive model considered in this study, the forecast equations for the level \(l_t \) and trend \(b_t \) by (\ref{w1}) - (\ref{w4}):

\begin{align}
	\hat{y}_{t+h} &= l_t + hb_t + s_{t+h-m(k+1)}  \label{w1}\\
	l_t &= \alpha (y_t - s_{t-m}) + (1 - \alpha)(l_{t-1} + b_{t-1}) \\
	b_t &= \beta (l_t - l_{t-1}) + (1 - \beta) b_{t-1} \\
	s_t &= \gamma (y_t - l_{t-1} - b_{t-1}) + (1 - \gamma) s_{t-m} \label{w4}
\end{align}
where \(\alpha \) is the level smoothing parameter ( \(0 < \alpha < 1 \) ), \( \beta \) is the trend smoothing parameter ( \( 0 < \beta < 1 \) ), \( \gamma \) is the seasonal smoothing parameter ( \( 0 < \gamma < 1 \) ), $h$ is the forecast horizon, \( m \) is the number of periods in a season, \( s_t \) is the seasonal component and \(k \) is the integer part of \( \frac{h-1}{m}\), \(y_t\) is the actual value at time \( t\), and \(\hat{y}_t \) is the forecasted value at time \(t\).
The ETS framework effectively models various patterns in time series data, making it suitable for forecasting monthly inflation rates. By adjusting the level, trend, and seasonal components, ETS models provide accurate and adaptable forecasts.

\subsection{Seasonal Auto Regressive Integrated Moving Averages (SARIMA) Model}
The introduction of the ARIMA model by \cite{box2015time} marked a significant advancement in time series analysis. However, to accommodate the unique challenges presented by seasonal variations, the SARIMA model was developed. This advancement allows for the explicit inclusion of seasonal components, thereby enhancing the model's applicability and effectiveness in scenarios where seasonal differentiation is crucial for making non-stationary time series data stationary \cite{agyemang2023time}. This makes SARIMA an indispensable tool in fields such as economics, finance, meteorology, and healthcare, where understanding and predicting seasonal trends can be critical to decision-making and strategy development. As postulated by \cite{box2015time,vanlaar2014evaluation}, the SARIMA model has four components and is represented in the study as follows:

\begin{itemize}
	\item The non-seasonal and seasonal Auto Regressive (AR) polynomial term of order $p$ and $P$ are expressed in (\ref{SAR1}) and (\ref{SAR2}) by:
	\begin{equation}
		\phi_p(B) = 1 - \phi_1 B - \phi_2 B^2 - \ldots - \phi_p B^p \label{SAR1}
			\end{equation}
			
	\begin{equation}		
		\Phi_P(B^s) = 1 - \Phi_1 B^s - \Phi_2 B^{2s} - \ldots - \Phi_P B^{Ps}
		\label{SAR2}
	\end{equation}
	
	\item The non-seasonal and seasonal Moving Average (MA) part of order $q$ and $Q$ are expressed in (\ref{SAR3}) and (\ref{SAR4}) by:
	\begin{equation}
	\theta_q(B)= 1 + \theta_1 B + \theta_2 B^2 + \ldots + \theta_q B^q
	\label{SAR3}
	\end{equation}

\begin{equation}
\Theta_Q(B^s) = 1 + \Theta_1 B^s + \Theta_2 B^{2s} + \ldots + \Theta_Q B^{Qs}	
\label{SAR4}
\end{equation}
	
	\item Non-seasonal differencing operator is of the order $d$ used to eliminate polynomial trends given in (\ref{SAR5}) by:
	\begin{equation}
		(1 - B)^d
		\label{SAR5}
	\end{equation}
	
	\item Seasonal differencing operator is the order of $D$ used to eliminate seasonal patterns is likewise given in  (\ref{SAR6}) by:
	\begin{equation}
		(1 - B^s)^D
		\label{SAR6}
	\end{equation}
\end{itemize}

\noindent
 The parameters $\phi$ and $\theta$ are the ordinary ARMA coefficients, $\Phi$ and $\Theta$ are the seasonal ARMA coefficients, $B$ is the backshift operator, whose effect on a time series $Y_t$ can be summarized in (\ref{SAR7}) by:
	\begin{equation}
		B^d Y_t = Y_{t-d}
			\label{SAR7}
	\end{equation}
\noindent
 Combining (\ref{SAR1}) - (\ref{SAR6}), the generalized form of the SARIMA$(p, d, q) \times (P, D, Q)_S$ model for a series $Y_t$ can be formulated in (\ref{SAR8}) as:
	\begin{equation}
		\phi_p(B) \Phi_P(B^s) (1 - B)^d (1 - B^s)^D Y_t = \theta_q(B) \Theta_Q(B^s) \epsilon_t
		\label{SAR8}
	\end{equation}
where $s$ is the length of the seasonality and  random errors $\epsilon_t$  are assumed as a white noise process.

\subsection{Model Assessment Metrics}
\noindent
In order to evaluate the competing models, Mean square error (MSE), mean absolute error (MAE) and geometric mean relative absolute error (GMRAE) and  Theil U1 Statistic $(\tau)$ are used. The computational formulae are given by (\ref{eq111}) - (\ref{eq444})

	\begin{equation}
		\text{MAE} = \frac{1}{n} \sum_{t=1}^{n} |y_t - \hat{y}_t|
		\label{eq111}
	\end{equation}
	
	\begin{equation}
		\text{MSE} = \frac{1}{n} \sum_{t=1}^{n} (y_t - \hat{y}_t)^2
		\label{eq222}
	\end{equation}

	\begin{equation}
		\text{GMRAE} = \exp \left( \frac{1}{n} \sum_{t=1}^{n} \ln \left( \left| \frac{y_t - \hat{y}_t}{y_t} \right| \right) \right)
		\label{eq333}
	\end{equation}

	 \begin{equation}
		\tau=\sqrt{\frac{1}{n}\sum_{t=1}^{n}\epsilon_t^2}\div \left(\sqrt{\frac{1}{n}\sum_{t=1}^{n}y_t^2}+\sqrt{\frac{1}{n}\sum_{t=1}^{n}\hat{y}_t^2}\right).
		\label{eq444}
	\end{equation}

\noindent
where $y_t$ are the actual values, $\hat{y}_t$ are the forecast values and $y_t-\hat{y}_t=\epsilon_t$ are the forecast errors. $0\le\tau\le1$, for $\tau\approxeq0$ implies good fit of model to data and $\tau\approxeq1$ implies poor fit of model to data.

\newpage
\section{Results and Discussion \label{Sec4}}
This section presents the results  and discussion of the analysis by the traditional ARIMA, ETS  and the six variants of deep learning models (Simple RNN, LSTM, GRU, BiLSTM, BiGRU and Transformer).

\subsection{Time Series Decomposition of Influenza A Cases}
\begin{figure}[!hbt]
	\centering
	\includegraphics[width=0.70\linewidth]{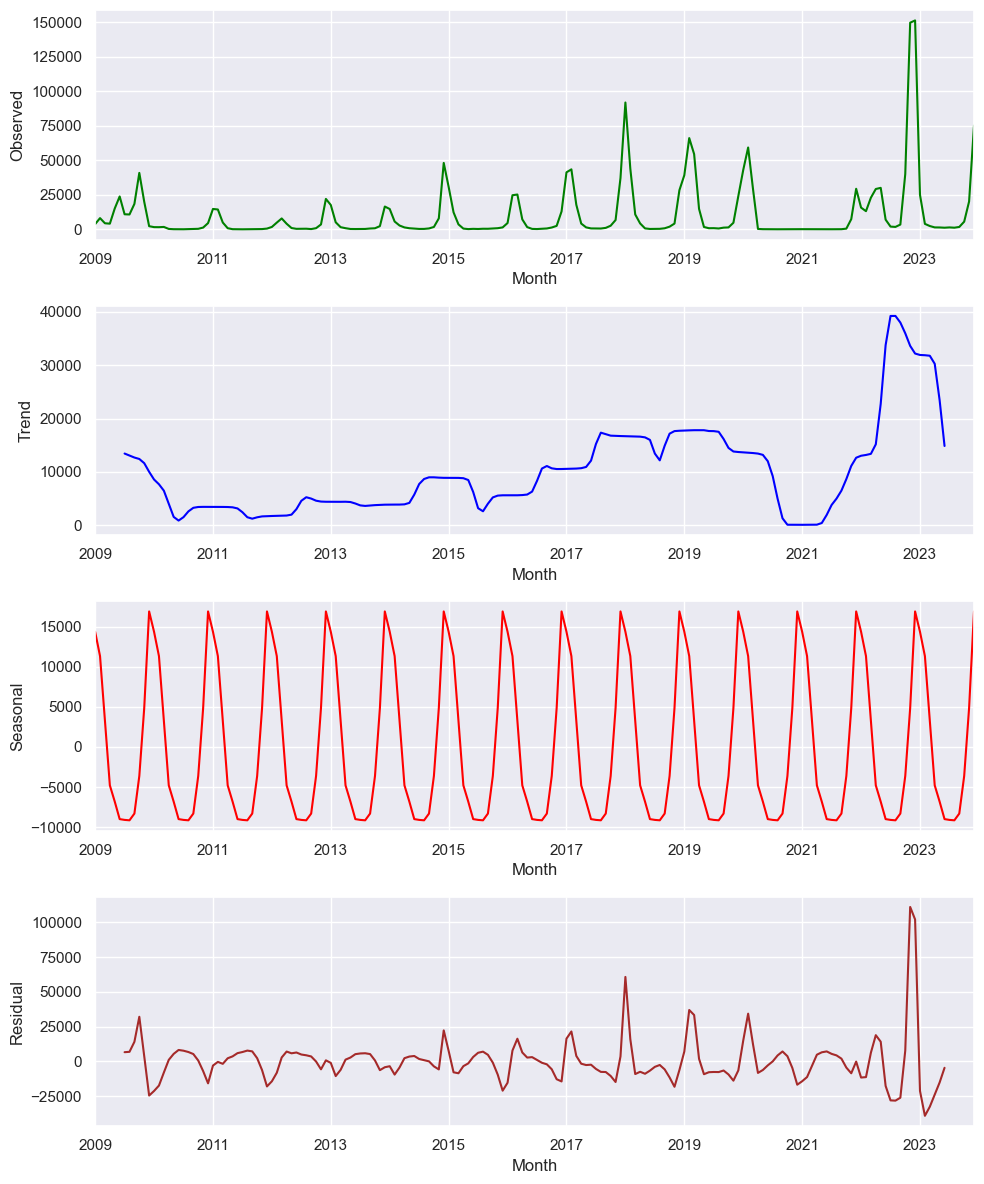}
	\caption{\textbf{Decomposition of the Influenza A series into trend, seasonality, and residuals}}
	\label{Seasonality}
\end{figure}

To analyze the seasonality and trend in the Influenza \textbf{A} cases data, we  employed a time series  additive decomposition method to decompose the series into its trend, seasonal, and residual (irregular) components as evident in Figure \ref{Seasonality}. This approach is essential in understanding how the Influenza \textbf{A} cases fluctuate over time within a year, highlighting any repeating patterns that occur on a seasonal basis.

\noindent
The trend component captures the long-term progression of Influenza \textbf{A} cases over time, showing how the level of Influenza \textbf{A} activity changes, disregarding the seasonal variations and irregular fluctuations. The seasonal component reveals the repeating patterns within each year, highlighting periods of higher or lower Influenza \textbf{A} activity. The seasonality component is particularly important for understanding when Influenza \textbf{A} cases tend to peak or decline during the year, which can be critical for public health planning and intervention strategies. The residuals represent the irregularities that cannot be explained by the trend or seasonal components. These could be random or non-systematic fluctuations that the model does not capture, possibly due to external factors (such as temperature, humidity, air pollution, etc) not included in the analysis.  From Figure \ref{Seasonality}, it is evident that Influenza \textbf{A} cases exhibit clear seasonal patterns, likely peaking in colder months when respiratory illnesses tend to spread more easily. The trend component suggests a variable trajectory of Influenza \textbf{A} cases over time, which could be influenced by a range of factors, including public health interventions, population immunity levels, and the emergence of new Influenza \textbf{A} strains. The residual component indicates that while the model captures much of the variation in Influenza \textbf{A} cases, some anomalies and irregularities remain unexplained, highlighting the complex nature of Influenza \textbf{A} dynamics and the potential influence of external factors. This decomposition provides critical insights into the behavior of Influenza \textbf{A} cases over time, emphasizing the importance of seasonality in understanding and managing Influenza \textbf{A} spread. Such insights can inform targeted public health responses and vaccination campaigns, aiming to mitigate the impact of seasonal Influenza \textbf{A} peaks.\\

\begin{figure}[!hbt]
	\centering
	\includegraphics[width=0.90\linewidth]{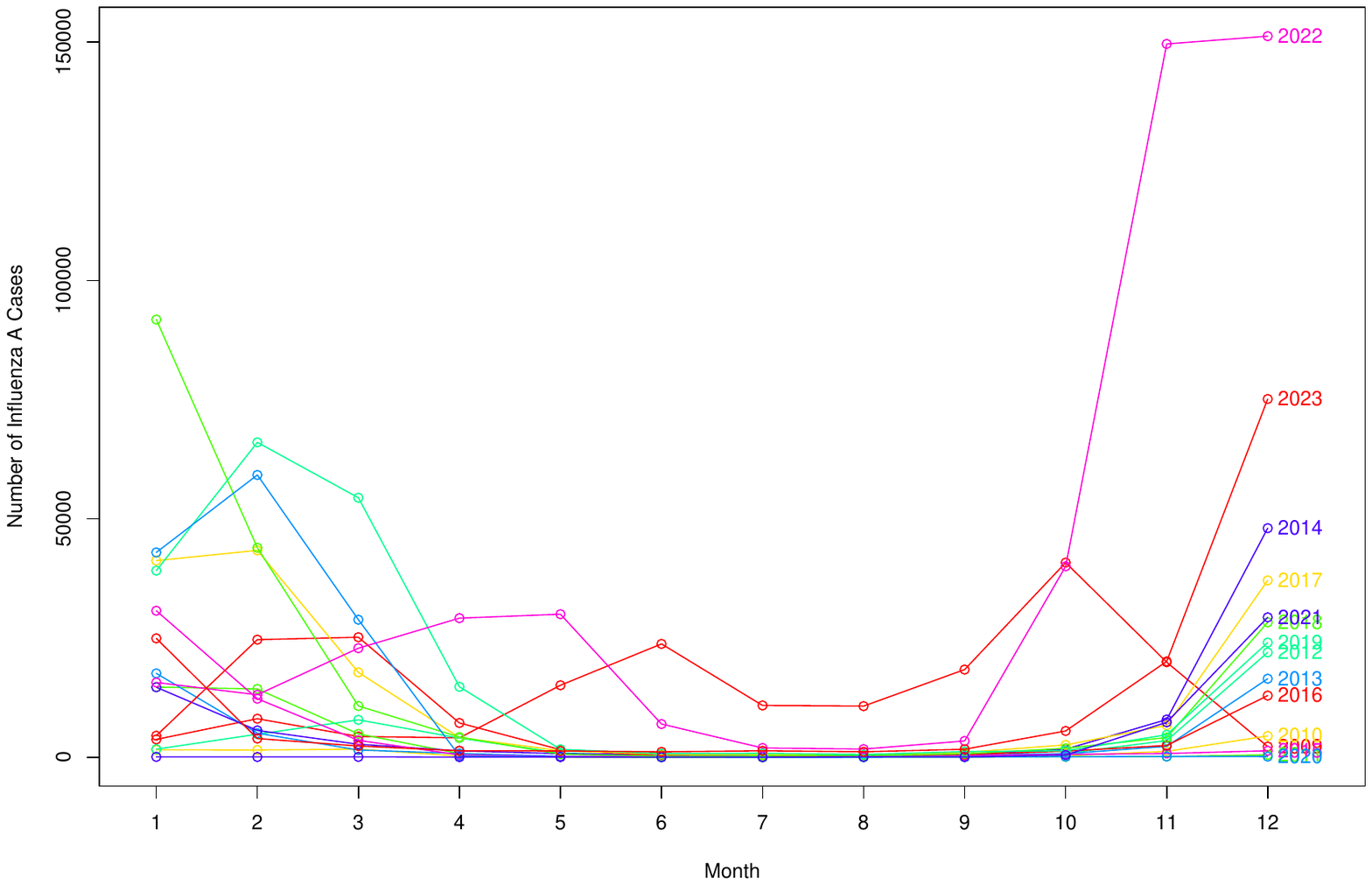}
	\caption{\textbf{Season plot of Influenza A Cases in USA }}
	\label{Seasonplot}
\end{figure}

\noindent
Figure \ref{Seasonplot} shows the number of Influenza \textbf{A} cases reported monthly from January, 2009 to December, 2023. Each line represents a different year, with the lines color-coded and labeled with the corresponding year for easy identification. From Figure \ref{Seasonplot}, we observe a seasonal pattern of Influenza \textbf{A} cases, with peaks generally occurring in the colder months towards the end of the year and into the early months of the following year. This seasonal trend is consistent with the expected behavior of Influenza \textbf{A} outbreaks, which typically peak during the colder months when people are more likely to be indoors and in close proximity to each other \citep{neumann2022seasonality}. Notably, the years 2022 and 2023 stand out due to their distinct patterns. In 2022, there's a dramatic spike in cases starting in November, reaching an exceptionally high peak by December, far exceeding the peaks of previous years. This surge suggests a significant outbreak or possibly reduced immunity in the population during that time. In contrast, 2023 shows a late start to the increase in cases, with a sharp rise beginning in October, indicating a possible delay in the onset of the seasonal outbreak or differences in virus transmission dynamics. Overall, Figure \ref{Seasonplot} illustrates the variability in Influenza \textbf{A} cases over time with emphasis on the seasonal nature of the disease and the exceptional outbreaks in the years 2022 and 2023.

\newpage
\noindent
The Mann Kendall Trend test was adopted in the study to test for possible existence of trend.

\begin{table}[!htbp]
	\centering
	\caption{\textbf{Mann Kendall Trend Test}}
	\label{Mann}
	\begin{tabular}{lcc}
		\toprule
		Test&Test Statistic &P-value\\ 
		\bottomrule
		\vspace{0.03in}
		Kendall Tau & 0.065&0.195\\ 
		\bottomrule
	\end{tabular} 
\end{table}

\noindent
From Table \ref{Mann}, the Mann Kendall test results suggest a very weak positive trend in the data, but the p-value indicates that this trend is not statistically significant at 5\% significance level. Therefore, we do not have sufficient evidence to conclude that there is a significant trend in the influenza \textbf{A} data.\\

\noindent
The Kruskall Wallis test of Seasonality was employed to confirm the existence of seasonality as claimed by Figure \ref{Seasonplot}. Since we suspected monthly seasonality, the data was divided into 12 groups, one for each month.  
\begin{table}[!htbp]
	\centering
	\caption{\textbf{Kruskall Wallis test of Seasonality}}
	\label{Krus}
	\begin{tabular}{lcc}
		\toprule
		Test&Test Statistic &P-value\\ 
		\bottomrule
		\vspace{0.03in}
		Kruskall Wallis  &65.091 &0.000\\ 
		\bottomrule
	\end{tabular} 
\end{table}

\noindent
From Table \ref{Krus}, the Kruskal-Wallis test indicates the presence of seasonal variations as the associated p-value (0.000) falls below the predetermined threshold of significance (0.05). Such a low p-value implies noteworthy differences among the groups, thereby warranting the existence of seasonality. This confirms the need for the usage of Seasonal ARIMA to model the Influenza \textbf{A} cases.

\begin{table}[!htbp]
	\centering
	\caption{\textbf{Descriptive Statistics of Influenza A Cases}}
	\label{Desr}
	\begin{tabular}{lccccc}
		\toprule
		Minimum&1st Quartile &Median&Mean&3rd Quartile&Maximum\\ 
		\bottomrule
		\vspace{0.03in}
		  12&344&1544&10286&10800&151234\\ 
		\bottomrule
	\end{tabular} 
\end{table}

\noindent
From Table \ref{Desr}, it is observed that the descriptive statistics for Influenza \textbf{A} cases reveal a broad spectrum of case counts, ranging from a minimum of 12 to a maximum of 151234. The data is skewed towards lower case counts, with the first quartile at 344. The median case count stands at 1544, highlighting that half of the cases fall below this mark. However, the mean case count significantly reduces to 10286, suggesting the presence of outliers or a few instances with extremely high case counts that elevate the average. This is further supported by the third quartile at 10800. The considerable gap between the maximum case count and the other descriptive measures underscores the existence of extreme values in the dataset, indicating that while the majority of Influenza \textbf{A} cases are of moderate count, there are rare but significant outbreaks that drastically exceed typical levels.

\newpage
\subsection{Analysis of the Holt-Winters Exponential Smoothing Model}
With the Influenza \textbf{A} series  exhibiting seasonality (established by the Kruskall-Wallis test), we first employ the Holt-Winters exponential smoothing to model the series.

\begin{figure}[!hbt]
	\centering
	\includegraphics[width=0.70\linewidth]{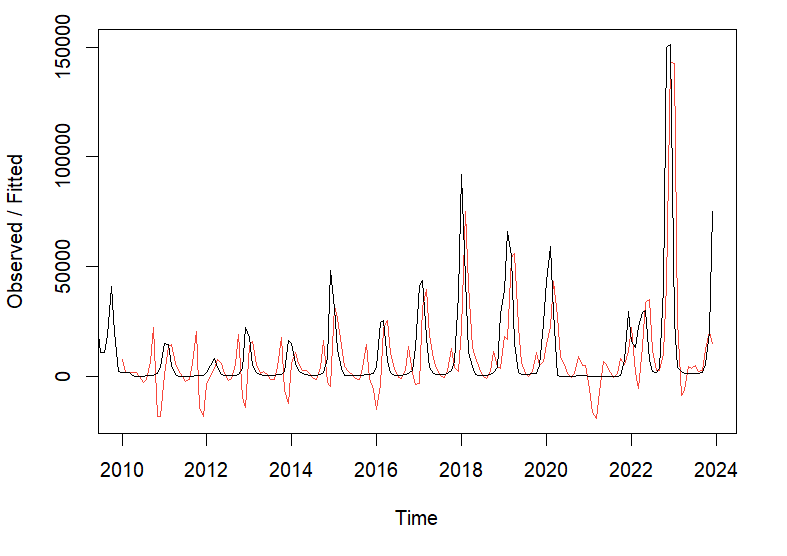}
	\caption{\textbf{Holt-Winters Filtering plot of Influenza A Cases in USA}}
	\label{Holt}
\end{figure}

\noindent
From Figure \ref{Holt}, Holt-Winters exponential method is very successful in predicting most of the seasonal peaks, which occur roughly during the last quarter of every year.

\begin{table}[!htbp]
	\centering
	\caption{\textbf{Holt-Winters exponential smoothing Parameters}}
	\label{Des}
\begin{tabular}{lcccccc}
	\toprule
	Para&Alpha &Beta&Gamma &Level &Slope&s1\\ 
	Value&0.887&0.000&1.000&44499.267&-1207.435&2758.229\\ \hline
	
	Para&s2 &s3&s4 &s5 &s6&s7\\ 
	Value&2066.236&-8292.859&-14998.426&-11105.509&-7941.158&-3359.326\\ \hline
	
	Para&s8&s9 &s10 &s11&s12&\\ 
	Value&-1313.436&1109.829&12015.127&27954.174&30647.733&\\ 
	\bottomrule
\end{tabular}

NB: Para; Parameter 
\end{table}

\noindent
The estimated values of alpha, beta and gamma are 0.887, 0.000, and 1.000 respectively as evident in Table \ref{Des}. The value of alpha ($\alpha=0.887$) is
relatively high, indicating that the estimate of the level component at the current time point is just based upon the very most recent observations.
The beta value is ($\beta=0.000$), which suggests that the slope $b$ of the trend component does not change over the time series and remains at its initial value. This is logical, given that while the level varies significantly throughout the time series, the slope $b$ of the trend component stays relatively constant. In contrast, the value of gamma ($\gamma=1.000$) is extremely high, indicating that the estimate of the seasonal component at the current time point is also based upon the very most recent observations just like the level component.

\subsubsection{Evaluation of Holt-Winters Exponential Smoothing model}
We investigate whether the predictive model can be improved upon by checking whether the in-sample forecast
errors show non-zero autocorrelations at lags 1-24 via the Ljung-Box test, residuals are normally distributed via the Lilliefors test and have constant variance via the Goldfeld-Quandt Homoscedasticity test.

\begin{table}[!htbp]
	\centering
	\caption{Ljung-Box, Lilliefors and Goldfeld-Quandt tests of ETS model}
	\begin{tabular}{lccccccccc}
		\toprule
		Time Series	& &  \multicolumn{2}{c}{\textbf{Ljung-Box}} && \multicolumn{2}{c}{\textbf{Lilliefors}}&&\multicolumn{2}{c}{\textbf{Goldfeld-Quandt}}\\ 
		
		\cmidrule{3-4} \cmidrule{6-7}\cmidrule{9-10}  \vspace{0.05in}
		&& $\chi^2$   & P-value &&$\chi^2$  & P-value  &&Statistic&P-value\\ 
		\hline
		Influenza \textbf{A}	&  & 111.581 & 0.000 & & 0.207 & 0.000 &&4.535&0.000 \\ 
		\bottomrule
	\end{tabular}
	\label{ETS}
\end{table}

\noindent
From Table \ref{ETS}, the p-value for Ljung-Box test is 0.000, indicating that there is sufficient evidence of non-zero autocorrelations at lags 1-24. This raises concerns about the model. We further check the normality and constant variance assumptions of the model. Also, from Table \ref{ETS}, with a p-value of 0.000, we can conclude that the residuals of the ETS model are not normally distributed. Likewise, from Table \ref{ETS}, we can infer that the variance of the residuals of ETS model  are not constant. Thus, with the violation of normality and constant variance assumptions, it is prudent to concede that the Holt-Winters model using Exponential smoothing can be improved upon. We resort to the ARIMA model in this regard.

\subsection{Analysis of the Auto Regressive Integrated Moving Averages (ARIMA) Model}
  \noindent
 The Kwiatkowski–Phillips–Schmidt–Shin Test (KPSS) test was employed to test for series stationarity. 
 
 \begin{table}[!htbp]
 	\centering
 	\caption{\textbf{KPSS test for Level Stationarity}}
 	\label{KPSS1}
 	\begin{tabular}{lccc}
 		\toprule
 		Test& Statistic &Truncation Lag  Parameter&P-value\\ 
 		\bottomrule
 		\vspace{0.1in}
 		KPSS &0.509&4&0.039\\ 
 		\bottomrule
 	\end{tabular} \\
 \end{table} 
 
\begin{figure}[!hbt]
 	\centering
 	\includegraphics[width=0.72\linewidth]{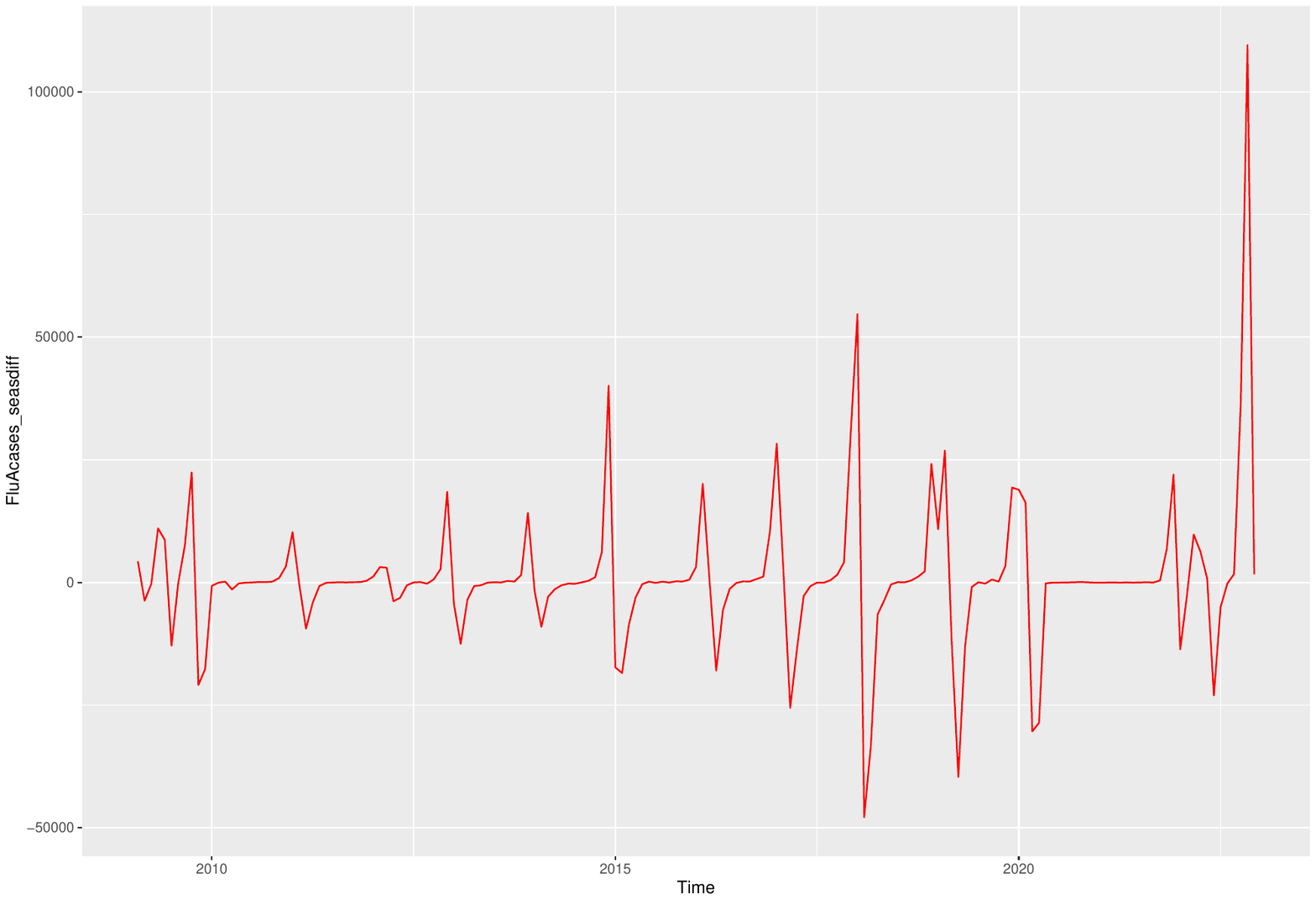}
 	\caption{\textbf{Single Exponential plot of Influenza A Cases in USA }}
 	\label{Diff}
 \end{figure}

 \noindent
 From Table \ref{KPSS1}, since the p-value (0.039), we reject the null hypothesis of series stationarity and conclude that the Influenza \textbf{A} series is not stationary. Hence, we subject the Influenza \textbf{A} cases series to a first-order differencing method to make it stationary. From Figure \ref{Diff}, the observed differenced data exhibits stationarity with constant mean and variance.

 \noindent
 The Kwiatkowski–Phillips–Schmidt–Shin Test (KPSS) test was employed to confirm  the series stationarity. 

\begin{table}[!htbp]
	\centering
	\caption{\textbf{KPSS test for Level Stationarity}}
	\label{KPSS}
	\begin{tabular}{lccc}
		\toprule
		Test& Statistic &Truncation Lag  Parameter&P-value\\ 
		\bottomrule
		\vspace{0.1in}
		KPSS &0.177&4&0.100\\ 
		\bottomrule
	\end{tabular} \\
\end{table} 

\noindent
We fail to reject the null hypothesis of stationarity and conclude that the first differenced Influenza \textbf{A} cases series is stationary since 0.100 $>$ 0.05 as evident in Table \ref{KPSS}.

\newpage
\subsubsection{Fitting  Suitable ARIMA model to Influenza A Cases}
After rigorous examination of both the initial first difference ACF and PACF plots, the following tentative ARIMA models were fitted to analyze Influenza \textbf{A} cases in the United States of America.

\begin{table}[!htbp]
	\centering
	\caption{\textbf{Possible tentative ARIMA models for Influenza A Cases}}
	\label{TabARMA}
	\begin{tabular}{lc}
		\toprule
		\textbf{Models}&\textbf{Evaluation metric (AIC)} \\ 
		\bottomrule
	ARIMA(0,1,0)(0,0,1)[12] &3650.484 \\
		ARIMA(0,1,0)(0,0,1)[12] with drift & 3652.038 \\
		ARIMA(0,1,0)(0,0,2)[12]  & 3651.999 \\
		ARIMA(0,1,0)(0,0,2)[12] with drift& 3653.567 \\
	ARIMA(0,1,0)(1,0,0)[12] &3652.349 \\
	ARIMA(0,1,0)(1,0,0)[12] with drift &3653.744 \\
		ARIMA(0,1,0)(1,0,1)[12] & 3654.075 \\
		ARIMA(0,1,0)(1,0,1)[12] with drift   & 3655.492 \\
\textbf{ARIMA(0,1,3)(0,0,1)[12]} & \textbf{3617.881} \\
		ARIMA(0,1,3)(0,0,1)[12] with drift & 3618.793 \\
	ARIMA(1,1,0)(0,0,1)[12] & 3647.634 \\
ARIMA(1,1,0)(0,0,2)[12] & 3649.582 \\
		ARIMA(2,1,0)(0,0,1)[12] & 3632.625 \\
		ARIMA(2,1,0)(0,0,2)[12] & 3634.598 \\
		ARIMA(2,1,0)(1,0,0)[12] with drift & 3638.043 \\
		ARIMA(2,1,0)(1,0,2)[12] & 3640.249 \\
		ARIMA(2,1,0)(1,0,2)[12] with drift & 3641.895\\
	ARIMA(3,1,0)(0,0,1)[12]& 3635.062 \\
	ARIMA(3,1,0)(0,0,1)[12] with drift & 3636.778 \\
		ARIMA(3,1,0)(0,0,2)[12] & 3637.127 \\
		\bottomrule
	\end{tabular} \\
	\textbf{Best model: ARIMA(0,1,3)(0,0,1)[12]}
\end{table}

\noindent From Table \ref{TabARMA}, it is evident that the \textbf{ARIMA(0,1,3)(0,0,1)[12]} is the ideal seasonal ARIMA model to forecast the underlying Influenza \textbf{A} cases since it has the least AIC value (3617.881) identified using the diagnostic of the ACF and PACF (with suggesting no additional AR terms) plots.

\begin{table}[!htbp]
	\centering
	\caption{\textbf{Final parameter estimates of Best model: ARIMA(0,1,3)(0,0,1)[12]}}
	\label{Param}
	\begin{tabular}{lcccc}
		\toprule
		Parameter&Coefficient &Standard Error&Z-value&P-value\\ 
		\bottomrule
		\vspace{0.02in}
		MA (1)&0.154& 0.077&2.004&0.045$^{***}$\\ 
		\vspace{0.02in}
		MA (2) &-0.599&0.093&-6.491&0.000$^{***}$\\ 
		\vspace{0.03in}
		MA(3) &-0.431& 0.098& -4.371 & 0.000$^{***}$ \\ 
		\vspace{0.03in}
		SMA (1) & 0.294& 0.106&2.774 &0.006$^{***}$\\ 
		\bottomrule
	\end{tabular} 
\end{table}
\noindent
From Table \ref{Param}, MA(1), MA (2), MA(3) and SMA (1) are statistically significant at the 5\% significance level.\\

\noindent
The general multiplicative Seasonal ARIMA model is thus given in the study by equation (\ref{SAR}) as:

	\begin{equation}
	\phi_0(B) \Phi_0(B^{12}) (1 - B)^1 (1 - B^{12})^0 Y_t = \theta_3(B) \Theta_0(B^{12}) \epsilon_t
	\label{SAR}
\end{equation}

\newpage
\subsubsection{Model Diagnostics of ARIMA(0,1,3)(0,0,1)[12]}
ARIMA(0,1,3)(0,0,1)[12] was subjected to model diagnostic to verify its suitability for forecasting.
\begin{table}[!htbp]
	\centering
	\caption{Ljung-Box, Lilliefors and Goldfeld-Quandt tests of SARIMA model}
	\begin{tabular}{lccccccccc}
		\toprule
		Time Series	& &  \multicolumn{2}{c}{\textbf{Ljung-Box}} && \multicolumn{2}{c}{\textbf{Lilliefors}}&&\multicolumn{2}{c}{\textbf{Goldfeld-Quandt}}\\ 
		
		\cmidrule{3-4} \cmidrule{6-7}\cmidrule{9-10}  \vspace{0.05in}
		&& $\chi^2$   & P-value &&$\chi^2$  & P-value  &&Statistic&P-value\\ 
		\hline
		Influenza \textbf{A}	&  & 13.405 & 0.340 & & 0.239 & 0.000 &&5.352&0.000 \\ 
		\bottomrule
	\end{tabular}
	\label{SARIMA}
\end{table}

\noindent
From Table \ref{SARIMA},with a p-value of 0.340, which exceeds the significance threshold of 0.05, we conclude that the residuals of ARIMA(0,1,3)(0,0,1)[12] model are independent and follow a white noise process. Likewise, the model displays no substantial lack of fit. At a 5\% significance level, since the p-value is less than 0.05 (thus, 0.000 $<$ 0.05), the residuals of the estimated ARIMA(0,1,3)(0,0,1)[12] model are not normally distributed as evident in Table \ref{SARIMA}. Likewise, from Table \ref{SARIMA}, it is observed that the variance residuals of the estimated ARIMA(0,1,3)(0,0,1)[12] are not constant (p$<$ 0.05). Thus, constant variance assumption is violated.

\subsubsection{Test of model adequacy of ARIMA(0,1,3)(0,0,1)[12]} 
\begin{figure}[!htbp]
	\centering
	\subfigure[Residual plot of ARIMA $\left(0,1,3\right)\times\left(0,0,1\right)_{12}$]
	{ \includegraphics[width=0.9\linewidth]{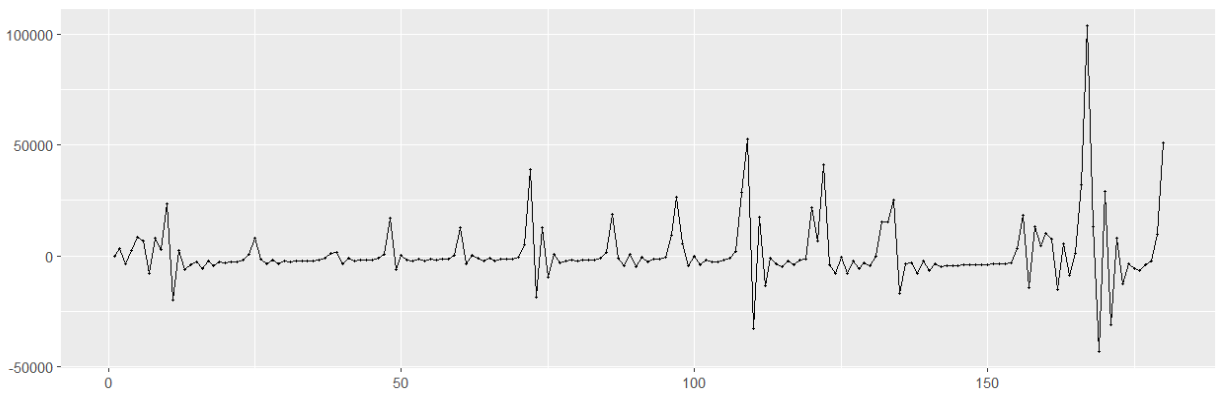}
		\label{E111}
	}\vfill
	\subfigure[ACF of ARIMA $\left(0,1,3\right)\times\left(0,0,1\right)_{12}$]
	{ \includegraphics[width=0.45\linewidth]{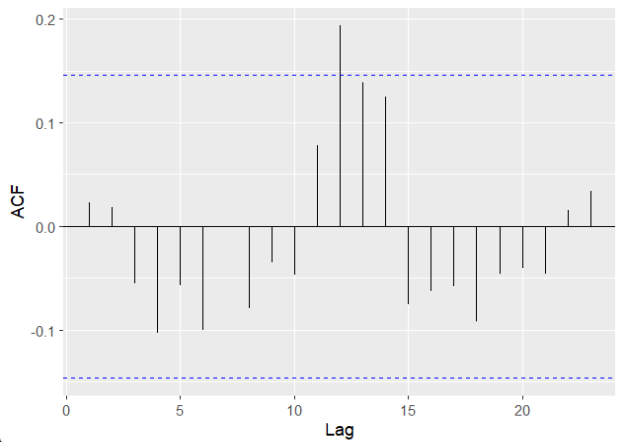}
		\label{E222}
	}\hfill
	\subfigure[PACF of ARIMA $\left(0,1,3\right)\times\left(0,0,1\right)_{12}$]
	{ \includegraphics[width=0.45\linewidth]{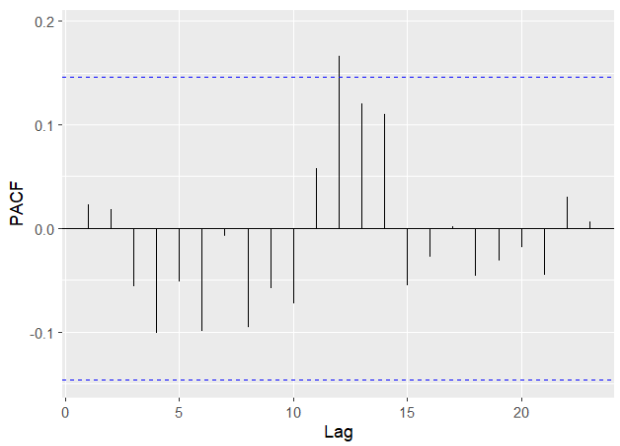}
		\label{E333}
	}
	\caption{\textbf{Plot of ACF and PACF plot of residuals of ARIMA $\left(0,1,3\right)\times\left(0,0,1\right)_{12}$ model}}
	\label{EE}
\end{figure}

\noindent
The limits of the 2$\sigma$ are given by $\pm Z_\frac{\alpha}{2}\times\frac{1}{\sqrt n}$.\\
\[-1.96\left(\frac{1}{\sqrt{168}}\right)\le r \le 1.96\left(\frac{1}{\sqrt{168}}\right)\]
This is simplified as $-0.151\le r \le 0.151$. Figure \ref{EE} shows two blue horizontal lines indicating this claim. It could be seen that the errors are within $\pm 0.151$ except  for lag 11 (for the ACF) and for lag 12 (for the PACF) as evident in Figures \ref{E222} and \ref{E333} respectively. The plot of the ACF and PACF shows no systematic structure, indicating that the residuals are purely random.

\subsubsection{Modified Ljung-Box test for larger lags}
To ascertain whether the proposed model is adequate for forecasting Influenza \textbf{A} cases, the modified Ljung-Box test (Box-Pierce test) for larger lags was employed .

\begin{table}[!htbp]
	\centering
	\caption{\textbf{Modified Ljung-Box (Box-Pierce test) for larger lags of ARIMA(0,1,3)(0,0,1)[12]}}
	\label{Lag}
	\begin{tabular}{lcc}
		\toprule
		Lags&Chi-Square Statistic &P-value\\ 
		\bottomrule
			12&12.455&0.409\\ 
		24&16.905&0.853\\ 
		36 &26.170&0.886\\ 
		48 &37.318&0.867\\ 
		60 &48.772&0.849\\ 
		72&53.829&0.946\\
		\bottomrule
	\end{tabular} 
\end{table}

\noindent
Upon examining Table \ref{Lag}, it becomes evident that the p-values across various lags significantly exceed the established 0.05 level of significance threshold. This leads to an insufficient foundation to refute our proposed ARIMA(0,1,3)(0,0,1)[12] model. Given this, and considering the model's satisfactory performance at lags 12, 24, 36, 48, 60, and 72, it is reasonable to infer that the model would likely remain adequate for larger lag periods as well. Consequently, the ARIMA(0,1,3)(0,0,1)[12] model demonstrates sufficient statistical reliability at the 0.05 significance threshold. \\

\noindent
It is to be noted that the GPR model employed by \cite{agyemang2025gaussian} using the same data identified several key anomalies (21 anomalies) in the Influenza \textbf{A} cases. These anomalies were characterized by substantial deviations from the predicted trend, suggesting periods of unexpected increases or decreases in Influenza \textbf{A} cases. See \cite{agyemang2024anomaly,agyemang2023unfolding,agyemang2023baseline,nortey2022bayesian} for further details on anomaly detection as anomalies could impact the results of public health data analysis. Even though the residuals of ARIMA(0,1,3)(0,0,1)[12] model are independent and suitable for larger lags, the violation of its normality and constant variance assumption raises concerns about the suitability of the model for forecasting future Influenza \textbf{A}. We thus switch our attention to deep learning methods such as Simple RNN, LSTM, GRU, BiLSTM, BiGRU and Transformer in search of a more suitable predictive model.

\subsection{Analysis using Deep Learning Architectures}
In this study, hyperparameter tuning of various deep learning networks (simple RNN, LSTM, GRU, BiLSTM, BiGRU, and Transformer) were performed in order to analyze Influenza \textbf{A} cases data. The objective is to reduce mean squared error (MSE) on the validation dataset and enhance generalization to unseen data. We tested different numbers of units (16, 32, 64, 128, 256, 512), activation functions (\texttt{relu}, \texttt{tanh}, \texttt{sigmoid}), learning rates (0.001, 0.01, 0.1), optimizers (\texttt{Adam}, \texttt{RMSprop}, \texttt{SGD}) and batch sizes (16, 32, 64, 128). Each configuration was built with a three-layer sequential model, including 30\% dropout to reduce overfitting and a  \texttt{TimeDistributed} dense layer for output. Training was performed with a 20\% validation split over 50 epochs for the models with less verbose output to focus on key metrics. The best performing hyperparameters were selected according to the minimum validation MSE. The complete hyperparameter tuning details of the deep-learning models considered in the study are presented in Table \ref{Tab3}.

\begin{table}[!htbp]
	\centering
	\caption{\textbf{Full Specification of the simple RNN, LSTM, GRU, BiLSTM and BiGRU models Architecture for Influenza A Forecasting after Hyperparameter Tuning}}
	\label{Tab3}
	\begin{tabular}{lccccc}
		\toprule
		Parameters&Simple RNN & LSTM&GRU&BiLSTSM&BiGRU\\ 
		\bottomrule
		\vspace{0.03in}
Number of Units &64 &64&64&64&64\\
Number of Layers  &4 &4&4&4&4\\
Batch Size  &32 &32&32&16&16\\
Activation Function  &Sigmoid  &Tanh&ReLu&Tanh&Tanh \\ 
Number of Epoch  &50 &50&50&50&50\\ 
Learning Rate  &0.001  &0.01&0.01&0.001&0.001\\ 
Recurrent Dropout  &None  &None&None&None&None\\ 
Optimizer  &RMsProp &Adam&Sgd&Adam&Adam\\ 
Dropout Rate &0.3  &0.3&0.3&0.3&0.3\\ 
Best Validation MSE  &0.0009&0.0006&0.0003&0.0004&0.0005\\ 
\bottomrule
	\end{tabular} 
\end{table}

\begin{figure}[!htbp]
	\centering
	\subfigure[Training and Test Predictions of LSTM]
	{ \includegraphics[width=0.80\linewidth]{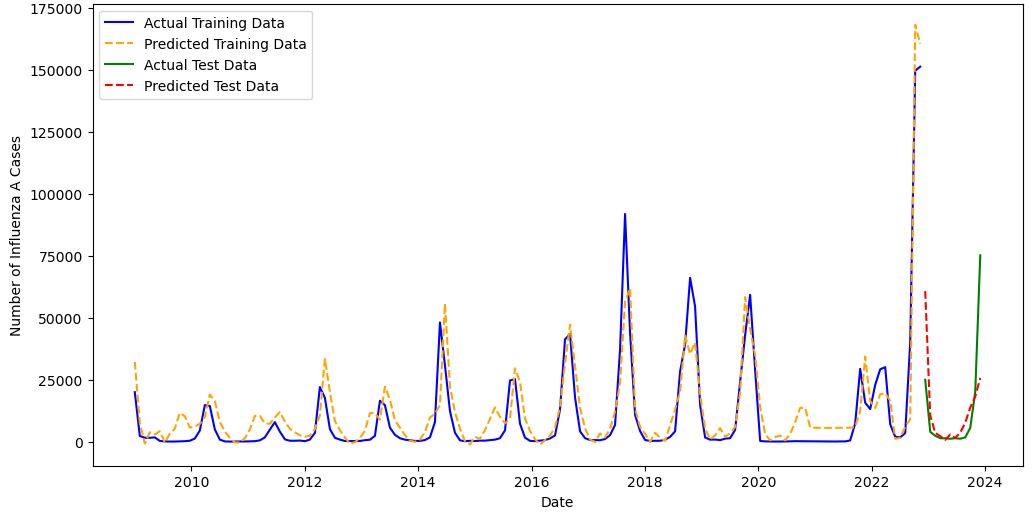}
		\label{L1}
	}\hfill
	\subfigure[Training and Testing Predictions of Simple RNN]
	{ \includegraphics[width=0.80\linewidth]{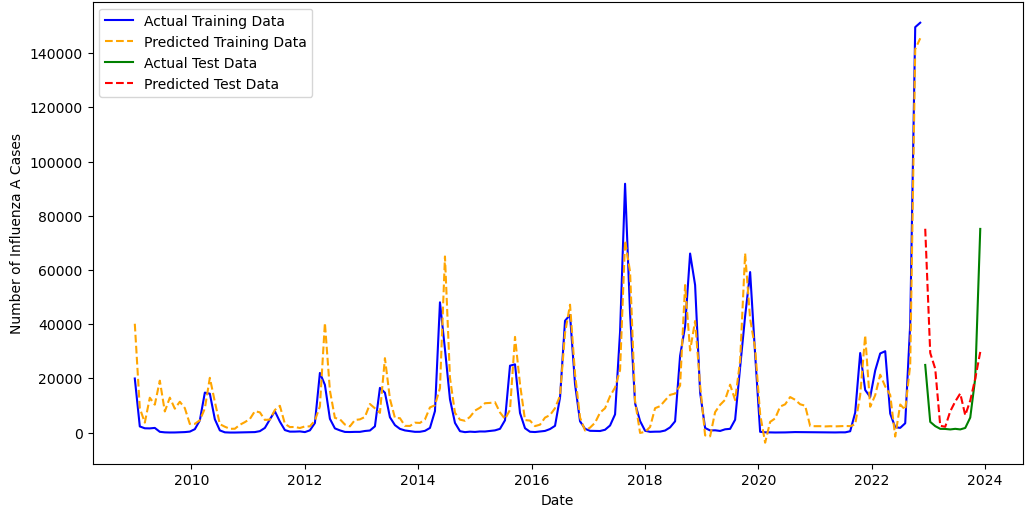}
		\label{S1}
	}\hfill
	\subfigure[Training and Test Predictions of GRU]
	{ \includegraphics[width=0.80\linewidth]{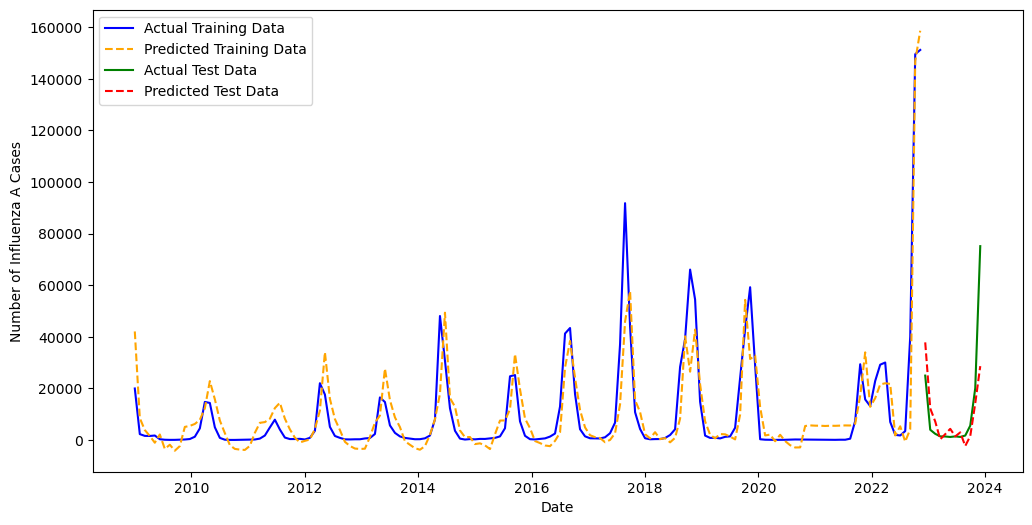}
		\label{G1}
	}
	\caption{\textbf{Train and Test Predictions of Influenza A (LSTM, RNN and GRU)}}
	\label{Inf}
\end{figure}

\newpage
\noindent
Figures \ref{Inf} gives the graphical representation of one of the runs of the training and testing predictions of LSTM, simple RNN and GRU models.

\subsection{Deployment of the Transformer Model}
The Transformer model was deployed 
by scaling the Influenza \textbf{A} cases between 0 and 1 for the deep learning model to perform optimally; in this study, the \texttt{MinMaxScaler} method was used. This set up ensure that the  model receives normalized data, reducing potential biases. Next, we converted the scaled time-series data into sequences that the Transformer model can process. Here, a function that takes a sequence length as input and constructs overlapping subsequences from the data was defined. Each sequence consists of a specified number of consecutive data points (e.g., 12 months of Influenza \textbf{A} cases) followed by the next point, which serves as the target value for the model to predict. The sequences were then transformed into PyTorch tensors as they are to be used in a Transformer model, which was built using PyTorch. Extensive Hyperparameter tuning resulting in the best hyperparameters (embedding dimension of 64, feed-forward hidden layer size of 128, dropout rate of 30\%, number of epochs of 50, learning rate of 0.001, number of heads - 3, number of layers - 4, batch size of 32, etc.) were used to build the model. The training was done using Adam optimizer and Mean Squared Error as the loss function over multiple epochs using \texttt{DataLoader}. This setup serves the dual purpose of maximizing learning while also calibrating the model weights to minimize prediction errors. Finally, the Transformer model's performance is evaluated by comparing predicted versus actual values and calculating error metrics like MSE, MAE, GMRAE and Theil U1 ($\tau$).\\

\begin{figure}[!htb]
	\centering
	\subfigure[Training and Test Predictions of Transformer]
	{ \includegraphics[width=0.81\linewidth]{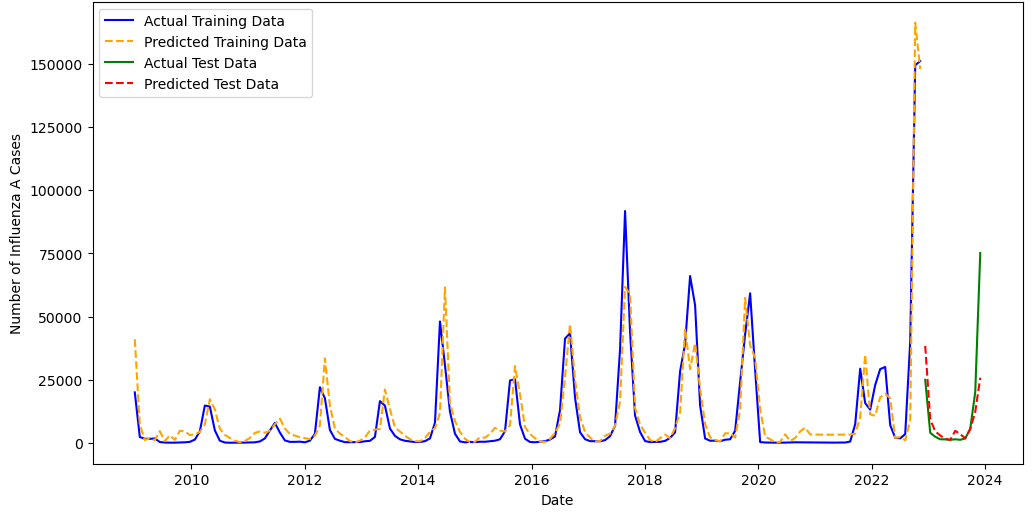}
		\label{T2}
	}\hfill
	\subfigure[Training and Test Predictions of BiLSTM]
	{ \includegraphics[width=0.81\linewidth]{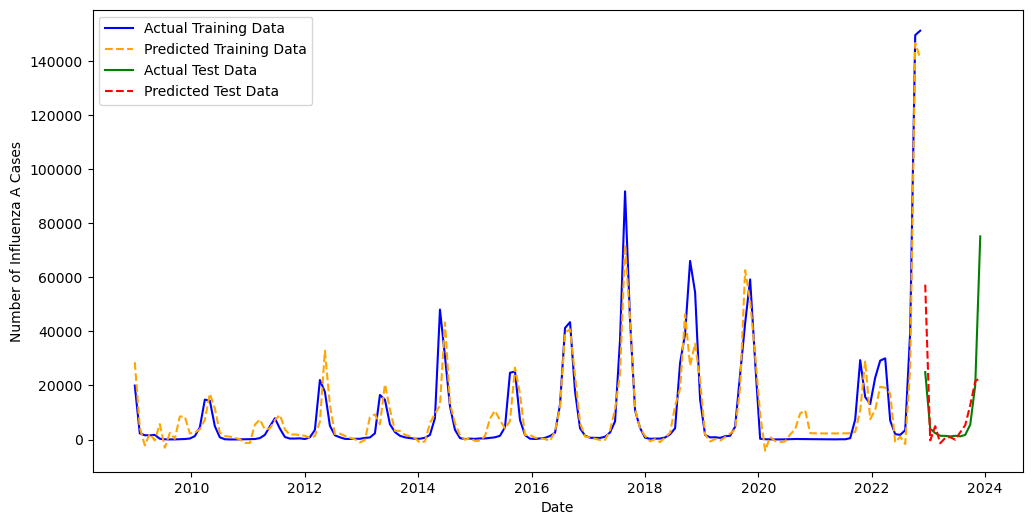}
		\label{L2}
	}\hfill
	\subfigure[Training and Test Predictions of BiGRU]
	{ \includegraphics[width=0.81\linewidth]{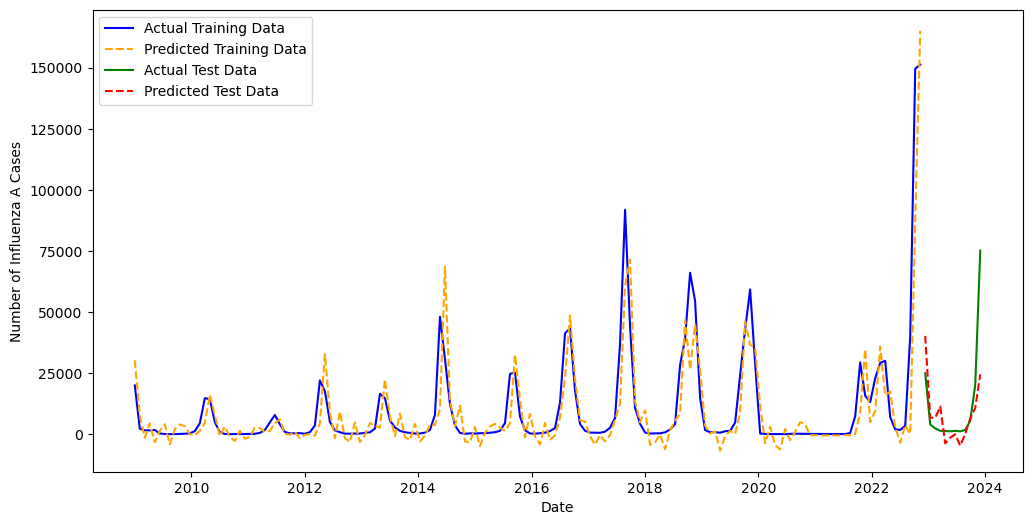}
		\label{G2}
	}
	\caption{\textbf{Train and Test Predictions of Influenza A  (Transformer, BiLSTM and BiGRU)}}
	\label{Inf1}
\end{figure}

\noindent
Figure \ref{Inf1} gives the graphical representation of one of the runs of the the training and testing predictions of Transformer, BiLSTM and BiGRU models. 

\begin{table}[htpb!]
	\centering
	\caption{\textbf{Simple RNN, LSTM, GRU, BiLSTM, BiGRU and Transformer Models Performance Evaluation Metrics of Influenza A Cases for Training Data}}
	\begin{tabular}{lcccc}
		\toprule
		\textbf{Models} & \textbf{Avg. MSE $\pm$  SE} & \textbf{Avg. MAE $\pm$  SE}&\textbf{Avg. GMRAE $\pm$  SE}&\textbf{Avg. $\tau$ $\pm$  SE} \\ 
		\midrule
		\vspace{0.06in}
		Simple RNN &0.0065$\pm$0.0006&0.0479$\pm$0.0067&1.0444$\pm$0.0068&0.4446$\pm$0.0241 \\ \vspace{0.06in}
		LSTM &0.0039$\pm$0.0008
		&0.0373$\pm$0.0056 &1.0340
		$\pm$0.0055&0.3002$\pm$0.0265
		\\ \vspace{0.06in}
		GRU&0.0044$\pm$0.0013&0.0417$\pm$0.0071&1.0386$\pm$0.0070&0.4013
		$\pm$0.0208\\  \vspace{0.06in}
	BiLSTM &0.0028$\pm$0.0002&0.0299
		$\pm$0.0028 &1.0268$\pm$0.0028
		&0.3113$\pm$0.0511\\  \vspace{0.06in}
		BiGRU&0.0029$\pm$0.0001&0.0335$\pm$0.0016 &1.0303$\pm$0.0016&0.2639$\pm$0.0053
		\\ \vspace{0.05in}
		\color{red}\textbf{Transformer} &\color{red}\textbf{0.0028}$\pm$\textbf{0.0001}&\color{red}\textbf{0.0298}$\pm$\textbf{0.0012 }&\color{red}\textbf{1.0267}$\pm$\textbf{0.0012}&\color{red}\textbf{0.2496}
		$\pm$\textbf{0.0071}\\
		\bottomrule
	\end{tabular}
	\label{TabComp1}
\end{table}

\begin{table}[htpb!]
	\centering
	\caption{\textbf{ARIMA, ETS, Simple RNN, LSTM, GRU, BiLSTM, BiGRU and Transformer Models Performance Evaluation Metrics of Influenza A Cases for Testing Data}}
	\begin{tabular}{lcccc}
		\toprule
		\textbf{Models} & \textbf{Avg. MSE $\pm$  SE} & \textbf{Avg. MAE $\pm$  SE}&\textbf{Avg. GMRAE $\pm$  SE}&\textbf{Avg. $\tau$ $\pm$  SE} \\ 
		\midrule
		\vspace{0.06in}
		ARIMA &$8.7465 \times 10^{8}$&29574.4270&20.09595&0.4660 \\ \vspace{0.06in}
		ETS &$1.2801\times10^{12}$&129848.7010&99.91031 &0.7871 \\ \vspace{0.06in}
		Simple RNN  &0.0602$\pm$0.0033&0.1346
		$\pm$0.0050&1.1085$\pm$0.0054
		&0.5513$\pm$0.0236 \\ \vspace{0.06in}
		LSTM &0.0545$\pm$0.0072&0.1243$\pm$0.0088 &1.1003$\pm$0.0086&0.4816
		$\pm$0.0349 \\ \vspace{0.06in}
		GRU& 0.0486$\pm$0.0096&0.1214$\pm$0.0079 &1.0976$\pm$0.0074
		&0.5465$\pm$0.0221\\ \vspace{0.06in}
		BiLSTM  &0.0502$\pm$0.0025&0.1129
		$\pm$0.0019&1.0864$\pm$0.0018
		&0.5005$\pm$0.0056\\  \vspace{0.06in}
		BiGRU&0.0519$\pm$0.0010&0.1148$\pm$0.0013 &1.0887$\pm$0.0012&0.4770$\pm$0.0215 \\ \vspace{0.06in}
		\color{red} \textbf{Transformer} &\color{red} \textbf{0.0433}$\pm$\textbf{0.0020}&\color{red}\textbf{0.1126}$\pm$\textbf{0.0016} &\color{red}\textbf{1.0877}$\pm$\textbf{0.0014}&\color{red}\textbf{0.4690}$\pm$\textbf{0.0184}\\
		\bottomrule
	\end{tabular}
	\label{TabComp2}
\end{table}

\begin{figure}[!hbt]
	\centering
	\includegraphics[width=0.90\linewidth]{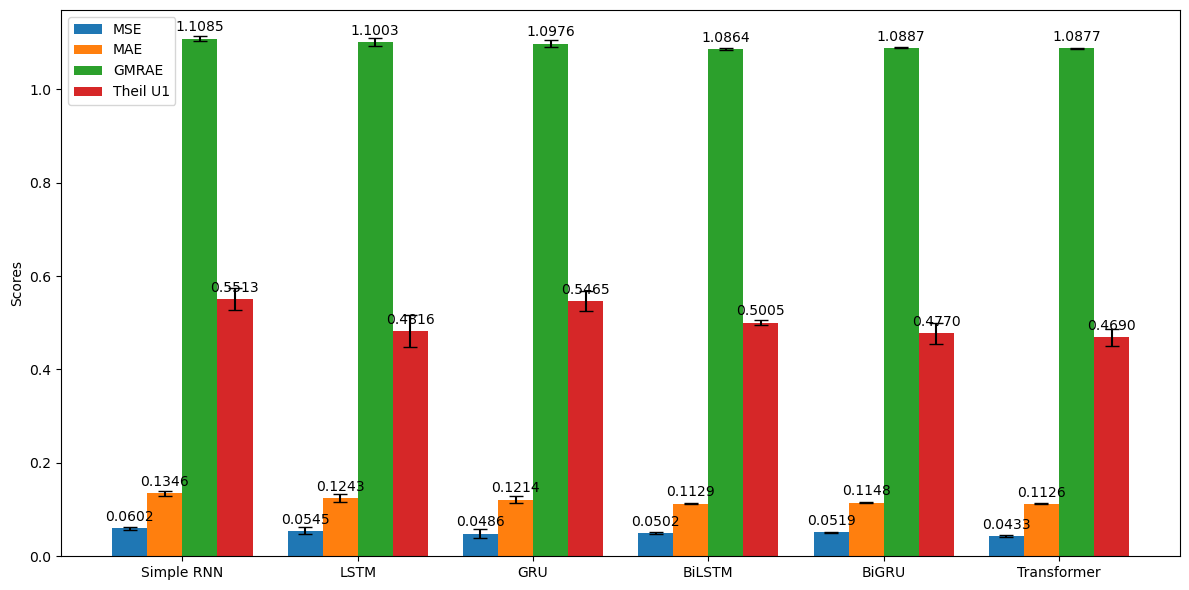}
	\caption{\textbf{Comparison of Error Metrics of the six deep learning models (Simple RNN LSTM, GRU, BiLSTM, BiGRU and Transformer)}}
	\label{ERC}
\end{figure}

\noindent
For the training data, in Table \ref{TabComp1}, it is observed that Transformer and BiLSTM show the best performance in terms of MSE and MAE, indicating lower prediction errors on average. While on average they have the same MSE'S (0.0028), their MAE values are also very close (0.0298 for Transformer vs. 0.0299 for BiLSTM) which suggests similar levels of accuracy in predicting the training data. Likewise, in the testing data, from Table \ref{TabComp2} and Figure \ref{ERC}, Transformer exhibits the lowest MSE and MAE of 0.0433 and 0.1126 respectively, suggesting it generalizes well on unseen data.

\subsection{Comparative Analysis with other Influenza time series models}
A Comparative analysis with numerous state-of-the-art studies, considering the methods applied, datasets used, and their best prediction errors, was conducted, as shown in Table \ref{TabAAA}.

\begin{table}[htpb]
	\centering
	\caption{\textbf{Comparison of RMSE and MAE of our study with other studies in literature that used Influenza \textbf{A} related dataset from 2021 to 2024}}
	\begin{tabular}{p{0.6in}p{2.2in}p{1.4in}p{2.05in}}
		\toprule
		\textbf{Research, Year} & \textbf{Methods used} & \textbf{Dataset used} & \textbf{Best Model Error Metrics} \\ \hline
\cite{tsan2022prediction}, 2022& SARIMAX, LSTM &CDC Influenza cases   &2174.80(RMSE) with SARIMAX\\ \hline
\cite{salem2024numerical}, 2024 & Fractional SEIR, ARIMA &Saudi Arabia Influenza cases   & 4.27(RMSE) and  0.36(MAE) with Fractional SEIR  \\ \hline
\cite{butkevych2024simulation}, 2024 & ARIMA &Ukraine Influenza cases & 37.5099(MAE) with ARIMA (10,1,11)\\ \hline
\cite{orang2024forecasting}, 2024&  SARIMA, NNAR&  Canada Influenza cases &2.57(RMSE) and  1.67(MAE) with NNAR(7, 1, 1) \\ \hline
\cite{mellor2023forecasting}, 2023 & GAM, ARIMA, Prophet &English sub-regions Influenza cases  &3.40(MAE) with GAM \\ \hline
\cite{arwaekaji2022forecasting}, 2022& Univariate SARIMA models & Thailand Influenza cases  &  0.46(RMSE) with SARIMA (1,0,1)(1,0,0) \\ \hline
\cite{li2024forecasting}, 2024& SARIMA, XGBoost, CNN, LSTM, CNN-LSTM & China Influenza cases  & 1.04(RMSE) and  0.46(MAE) with CNN-LSTM\\ \hline
\cite{seba2024time}, 2024 &SARIMA, ETS, BATS, RNN  & Algeria Influenza cases  & 71.34(RMSE) with RNN\\ \hline
\cite{zhu2022lasso}, 2022 &LASSO, ARIMA, RF, OLS, LSTM, Naive  &  China Influenza cases &  0.89(RMSE) and  1.04(MAE) with LASSO\\ \hline
\cite{alzahrani2024improving}, 2024& XGBoost, ARIMA, SARIMA & Saudi Arabia Influenza cases  & 274.82(RMSE) and  192.61(MAE) with XGBoost\\ \hline
\cite{du2022exploration}, 2022& Univariate SARIMA models & China Influenza cases  & 102.58(RMSE) and  53.82(MAE) with SARIMAX (1,0,0)(1,0,1)\\ \hline
\cite{zhao2023study}, 2023 & SARIMA, SARIMA-LSTM, SSA-SARIMA-LSTM & China Influenza cases  & 0.33(RMSE) and  0.23(MAE) with SSA-SARIMA-LSTM \\ \hline
\cite{kara2021multi}, 2021 & RF, SVR, ARIMA, FCNN, Hybrid GA-LSTM & USA Influenza cases  & 5166(RMSE) and  3173(MAE) with Hybrid GA-LSTM\\ \hline
\cite{olukanmi2021utilizing}, 2021 & LSTM, FNN, MLR, EN, SVM, SARIMA &  South Africa Influenza like Illness  & 10.54(RMSE) and  7.33(MAE) with FNN\\ \hline
\cite{poonawala2021novel}, 2021&AR, ARIMA, ETS, TBATS, Prophet, RF,  RF-REACH & New Zealand Influenza cases & 4.18(RMSE) with RF-REACH and  3.21(MAE) with AR \\ \hline
\cite{zhao2023new}, 2023& SARIMA, ETS, SARIMA-ETS-SVR &  China Influenza cases & 7.01(RMSE) and  1.62(MAE) with SARIMA-ETS-SVR\\ \hline
\cite{chhabra2024sustainable}, 2024& FB-Prophet, ARIMA, Holts Winter, Polynomial Regression & CSSE Influenza cases  & 14139.51(RMSE) with FB-Prophet\\ \hline
\cite{chen2024prediction}, 2024& SARIMA, Holts Winter, FB, XGBoost & China Influenza cases & 435.82 (RMSE) and 128.69 (MAE) with XGBoost\\ \hline
\cite{shen2024forecasting}, 2024&PatchTST, WhiteTST, TimesNet, ETSformer, LightTS, DLinear, Non-Stationary,LSTM  & CDC Influenza cases & 1.20 (RMSE) and 0.75 (MAE) with WhiteTST\\ \hline
\textbf{Current Study} &ARIMA, ETS, RNN, LSTM, GRU, BiLSTM, BiGRU, Transformer  &USA Influenza cases &0.21(RMSE) and 0.11(MAE) with Transformer  \\ 
\bottomrule
	\end{tabular}
	\label{TabAAA}
NNAR: Neural Network Autoregressive; GAM: Generalised Additive model; SSA: Singular spectrum analysis; FCNN: Fully Connected Neural Network; EN: Elastic Net; REACH: Randomized Ensembles of Auto-regression Chains; GA: Genetic Algorithm; WhiteTST: White-box Time Series Transformer.
\end{table}

\newpage
\section{Conclusions and Recommendations \label{Sec5}}
%done
Influenza \textbf{A} accounts for between 290,000 to 650,000 respiratory deaths each year, although this number is a marked decrease from past decades due to improved sanitation, healthcare practices, and vaccination approaches \citep{WHO2018Influenza}. Despite these advancements, Influenza \textbf{A} remains a major global health challenge, and this can be attributed to two key reasons. First, inadequate healthcare access and low Influenza vaccination rates persist across various global regions \cite{dini2018influenza}, increasing the chances of Influenza related mortality. Due to these gaps in healthcare and preventative measures, many individuals around the world continue to be vulnerable to the flu. Second, the demographic change towards older populations in many countries increases the overall risk of serious Influenza outbreaks \cite{bansal2010shifting}. This trend is of particular concern in lower-income countries, where the growing average age may increase the disease's burden in upcoming years, as older populations are generally more susceptible to serious complications from Influenza. To address these challenges, the global community could draw on historical strategies that have successfully mitigated Influenza's impact \citep{zachreson2020interfering}. Improving rates of vaccination, not only for Influenza but also for other communicable diseases, will reduce the severity of infections across the board. In addition, worldwide initiatives should address sanitation policies while preventing the spread of Influenza, as well as increasing access to healthcare for individuals regardless of location and income. Such steps are critical for minimizing the global impact of Influenza and protecting global health from future outbreaks.
\\

\noindent
%done
This study offers a comprehensive comparison of traditional and modern deep learning forecasting models for Influenza \textbf{A} cases, utilizing two versions of traditional models (ARIMA, ETS) along with six versions of deep learning models (Simple RNN, LSTM, GRU, BiLSTM, BiGRU, and Transformer). The historical Influenza \textbf{A} data is utilized to evaluate the accuracy of the prediction of each model in the analysis. The seasonal nature of Influenza \textbf{A} with peaks in colder months highlights the need for robust forecasting models. Although the ARIMA model handles seasonality well, in our case, the performance of ARIMA was not as strong as that of other deep learning architectures, especially state-of-the-art Transformer and GRU models that can better handle time-dependent data. The deep learning models show significant gain in forecasting accuracy compared to traditional ARIMA and ETS methods, as they are superior in capturing long-term dependencies. The results of this study reveal a clear superiority of all the deep learning models, especially the state-of-the-art Transformer with respective average testing MSE and MAE of $0.0433 \pm 0.0020$ and $0.1126 \pm 0.0016$ for capturing the temporal complexities associated with Influenza \textbf{A} data, outperforming well known traditional baseline ARIMA and ETS models. This advantage corroborates with existing state-of-the-art literature regarding the performance of deep learning models with respect to sequential and time-series data \citep{torres2021deep}.\\

\noindent
%done
It was observed that the Simple RNN, while outperforming the traditional ARIMA (0,1,3)(0,0,1)[12] and ETS models, fell short compared to state-of-the-art deep learning models like Transformer, LSTM, BiLSTM, GRU, and BiGRU counterparts, likely due to its limitations in capturing long term dependencies. These findings of this study provide evidence that state-of-the-art deep learning architectures can enhance predictive modeling for infectious diseases and indicate a more general trend toward using deep learning methods to enhance public health forecasting and intervention planning strategies. The superiority of state-of-the-art deep learning models can be attributed to processing sequential information better and being better adapted to the temporal dynamics that infectious disease data exhibit \citep{jin2020urban}.  Nonetheless, it is important to acknowledge the computational complexity and data prerequisites that come with using deep learning architectures, which can constrain their usage. However, regardless of these challenges, the evidence based predictive performance of deep learning frameworks, including state-of-the-art Transformer are well documented in the literature. Public health strategies could be significantly impacted by shifting to more complex forecasting techniques, ultimately leading to timely intervention resulting in a decrease in Influenza \textbf{A} morbidity and mortality. The study therefore recommended the incorporation of  Transformer models into real time forecasting and preparedness systems at an epidemic level, integrated into existing surveillance systems.

\subsection*{Author contributions}
\noindent
The authors declare to have contributed equally to the manuscript. All authors read and approved of the final manuscript.

\subsection*{Conflict of Interest}
\noindent
The authors declare that there are no conflicts of interest.

\subsection*{Ethical Statement}
\noindent
Not applicable as the study made use of a secondary data which is publicly accessible.

\subsection*{Acknowledgment}
\noindent
The first and corresponding author acknowledges the enormous support of the University of Texas Rio Grande Valley (UTRGV) Presidential Research Fellowship fund. The authors also thank Professor Tamer Oraby, Professor George Yanev and Professor Zhuanzhuan Ma for their guidance and feedback, which greatly shaped this work. We also thank the reviewers for their valuable comments and suggestions. 

\renewcommand{\bibname}{References}
\bibliographystyle{unsrt}
%\nocite{*}
%\bibpunct{(}{)}{,}{a}{}{;}
\bibliography{REFERENCES}
\end{document}